\documentclass{article}

\usepackage{microtype}
\usepackage{graphicx}
\usepackage{subfigure}
\usepackage{bm}
\usepackage{amsmath}
\usepackage{xcolor}
\usepackage{booktabs} % for professional tables
\usepackage[accepted]{icml2020}

\begin{document}

\twocolumn[
\icmltitle{From Quantized DNNs to Quantizable DNNs}

\begin{icmlauthorlist}
\icmlauthor{Kunyuan Du}{to}
\icmlauthor{Ya Zhang}{to}
\icmlauthor{Haibing Guan}{to}
\end{icmlauthorlist}

\icmlaffiliation{to}{Shanghai Jiao Tong University}
\icmlcorrespondingauthor{Kunyuan Du}{dukunyuan@sjtu.edu.cn}
\icmlkeywords{Machine Learning, ICML}

\vskip 0.3in
]



\printAffiliationsAndNotice{}  % leave blank if no need to mention equal contribution

\begin{abstract}

This paper proposes \emph{Quantizable DNNs}, a special type of DNNs that can flexibly quantize its bit-width (denoted as `bit modes' thereafter) during execution without further re-training. To simultaneously optimize for all bit modes, a combinational loss of all bit modes is proposed, which enforces consistent predictions ranging from low-bit mode to 32-bit mode. This \emph{Consistency-based Loss} may also be viewed as certain form of regularization during training. Because outputs of matrix multiplication in different bit modes have different distributions, we introduce \emph{Bit-Specific Batch Normalization} so as to reduce conflicts among different bit modes. Experiments on CIFAR100 and ImageNet have shown that compared to quantized DNNs, Quantizable DNNs not only have much better flexibility, but also achieve even higher classification accuracy. Ablation studies further verify that the regularization through the  consistency-based loss indeed improves the model's generalization performance.

\end{abstract}

\section{Introduction}

\begin{figure*}[htbp]
\vskip 0.2in
\begin{center}
\centerline{\includegraphics[width=18cm]{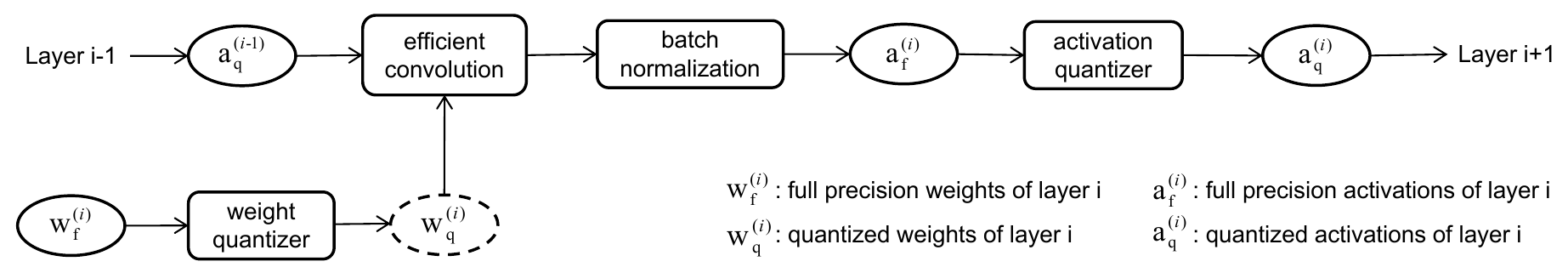}}
\caption{Quantization-aware training. }
\label{qat}
\end{center}
\vskip -0.2in
\end{figure*}

With increasing complexity of Deep Neural Networks (DNNs), great challenges are faced when deploying DNN models to mobile and embedded devices. As a result, model compression and acceleration have received more and more attention in the machine learning community. An important line of research is quantized DNNs, which convert both weights and activations to discrete space. Due to the reduction in bit-width, quantized DNNs have much smaller model size and can be inferenced with high-efficiency fixed-point computation for acceleration. However, when directly quantizing DNNs to $\leq$ 4 bits, significant accuracy degradation occurs. To alleviate this problem,  quantization-aware training (Fig.~\ref{qat}), which simulates the quantization effect with certain bit-width during training and allows the model to adapt to the quantization noise, is widely adopted ~\cite{hubara2016binarized, zhang2018lq, jung2019learning}.

\begin{figure}[t]
%\vskip 0.2in
\begin{center}
\centerline{\includegraphics[width=8.5cm]{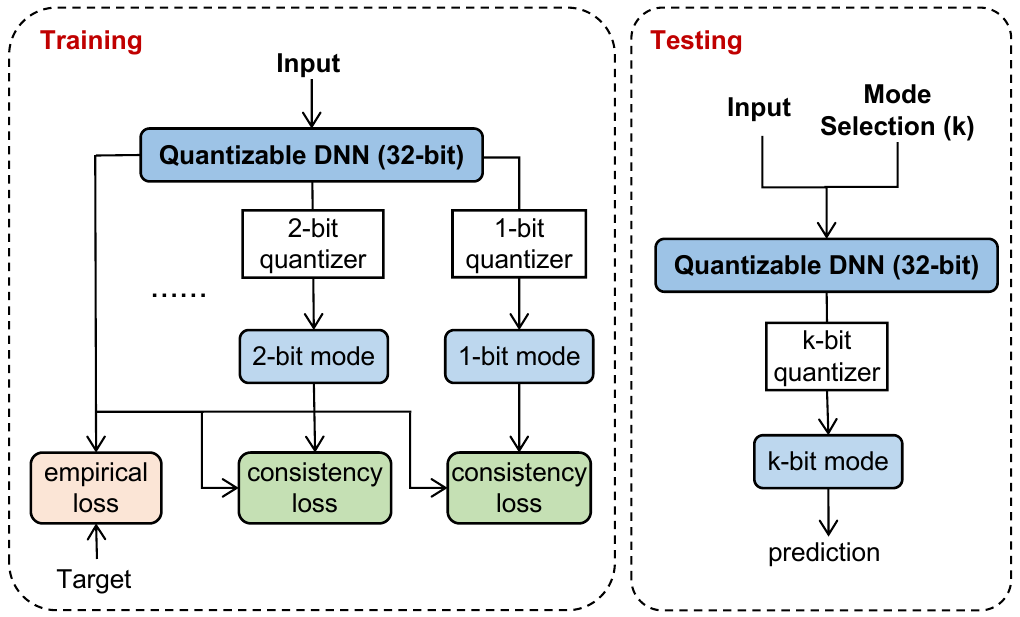}}
\caption{Illustration of Quantizable DNNs. Quantizable DNNs have approximately the same number of parameters as $k$-bit mode.}
\label{quantizable_dnn}
\end{center}
\vskip -0.2in
\end{figure}

\begin{table}[b]
\vspace{-0.85cm}
\centering
\caption{Supported bit-widths on different devices.}
\smallskip
\begin{tabular}{ccccccc}
  \toprule
   & Tesla T4 & Watt A1 & FPGA-based   \\
  \midrule
  Bit-widths & 32, 16, 8, 4 & 4, 3, 2, 1 & custom\\
  \bottomrule
 \end{tabular}
\label{device}
\end{table}

In real-world scenarios, different bit-widths may be supported by different devices, as shown in Table~\ref{device}. Either to easily deploy models to different devices, or to dynamic accuracy-efficiency trade-offs on the same device, it is desired to flexibly adjust the bit-width of quantized DNNs. However, it is usually non-trivial to re-configure the bit-width of quantized DNNs, because further quantization-aware re-training is required in order to maintain model accuracy. 

We thus propose `Quantizable' DNNs, a special type of quantized DNNs that can flexibly adjust its bit-width on the fly, i.e. turning on different bit mode by simply applying different quantization precision.
Quantizable DNNs target to pursue a single optimal set of convolutional kernels and fully-connected weights so that different bit modes achieve high accuracy at the same time.

Treating the optimization of different bit modes as a set of related sub-tasks, a multi-task framework is adopted to optimize the Quantizable DNNs. Fig. \ref{quantizable_dnn} provides an illustration for the Quantizable DNNs. The 32-bit (full-precision) mode serves as the `parent' model for lower bit modes. For 32-bit mode, the loss function is simply the empirical loss as that of individual quantized DNNs. To optimize the lower bit modes, a \emph{consistency loss} is further introduced to encourage different bit modes to produce consistent predictions to 32-bit mode. With the consistency loss, lower bit modes are implicitly guided by 32-bit mode to better fit the training set, since the former suffer more from noise in gradients~\cite{yin2019understanding} and are easily trapped in poor local minima~\cite{zhuang2018towards} during training. 
On the other hand, lower bit modes can be considered to provide a certain form of regularization to 32-bit mode through the consistency loss, since the lower bit modes are expected to emphasize on more critical information rather than the redundant details.
The lower-bit regularization is also similar to the well-known `Dropout' technique~\cite{JMLR:v15:srivastava14a}, by removing less significant bits during training. 

Another challenge faced with optimizing the Quantizable DNNs is that outputs of convolutional operation in different bit modes have different distributions, which makes it difficult to properly normalize feature maps from all bit modes with a shared Batch Normalizations~\cite{ioffe2015batch}. Inspired by~\cite{li2018adaptive, yu2018slimmable}, we introduce \emph{Bit-Specific Batch Normalization} to alleviate this problem, which assigns a separate Batch Normalization to each bit mode. The \emph{Bit-Specific Batch Normalization} introduces only negligible additional weights and hence Quantizable DNNs still have approximately the same number of parameters as quantized DNNs.

To validate the effectiveness of Quantizable DNNs, we conduct experiments on two widely used benchmark data sets, Cifar100~\cite{krizhevsky2009learning} and ImageNet~\cite{deng2009imagenet}. Compared with quantized DNNs, Quantizable DNNs not only enable the on-the-fly dynamic adjustment of mode bit-widths, but also achieve higher classification accuracy with the mutual regularization among different bit-widths. The main contributions of this paper are summarized as follows. 
\begin{itemize}
    \item We design the Quantizable DNN, which is the first DNN model that dynamically adjusts its bit-widths (bit modes) on the fly.
    \item We explore a multi-task co-regularization framework with \emph{consistency loss}, which enables the lower bit modes and the 32 bit mode to mutually promote each other during training.
    \item We propose \emph{Bit-Specific Batch Normalization} to alleviate the distribution difference among different bit modes, so that the same parent model may be shared.

\end{itemize}

\section{Related Work}

\subsection{Quantized DNNs}

For smaller model size and higher computational efficiency, both weights and activations of quantized DNNs lie in discrete spaces. Considering whether or not a method needs further re-training, it can be divided into post quantization~\cite{jacob2018quantization,zhao2019improving, nagel2019data} and quantization-aware training quantization~\cite{hubara2016binarized, zhou2016dorefa, lin2017towards, zhang2018lq, wang2019learning}. Most post quantization methods are limited to 8-bit values, and significant performance degradation occurs for $\leq 4$ bit quantization. To solve this problem, quantization-aware training can be applied, which considers quantization noise during training, as is shown in Fig.~\ref{qat}. During feed-forward calculation, full precision weights $w_{f}^{(i)}$ are first quantized to low-bit precision $w_{q}^{(i)}$ before matrix multiplication. And the activations $a_{f}^{(i)}$ are also quantized to $a_{q}^{(i)}$ before being fed into the next layer. During back-propagation, $w_{f}^{(i)}$ is updated rather than $w_{q}^{(i)}$. However, such training process is bit-specific, and converged quantized DNN cannot directly switch to other bit-widths. 

In this paper, Quantizable DNNs are implemented based on quantized DNNs, thus careful selection for the base model is needed. We choose Dorefa-net~\cite{zhou2016dorefa} for the following reasons. Firstly, unlike ~\cite{zhang2018lq}, it adopts uniform quantization scheme, which makes it much easier to deploy in various embedded products, e.g. Megvii. Secondly, it is applicable to common network architectures, while~\cite{lin2017towards, wang2019learning} requires specially-designed structure. Below we give a brief introduction to Dorefa-Net. In Dorefa-Net, the $k$-bit weight quantizer and $k$-bit activation quantizer are defined as Eq.(\ref{weight_quantizer}) and Eq.(\ref{activation_quantizer}) respectively. 
\begin{equation}
w_{q} = 2Q_{k}(\frac{\tanh(w_{f})}{2\max(|\tanh(w_{f})|)} + \frac{1}{2})-1, 
\label{weight_quantizer}  
\end{equation}
\begin{equation}
a_{q} = Q_{k}(clamp(a_{f}, 0, 1)), 
\label{activation_quantizer}  
\end{equation}
where $Q_{k}(\cdot)$ is the $k$ bit pre-defined quantizer which maps a real number $r \in [0,1]$ to a discrete value $q \in \{\frac{i}{2^{k}-1} | \ 0\leq i \leq 2^{k}-1, i \in N \}$, which is formulated as Eq.(\ref{quantizer}). 
\begin{equation}
q = Q_{k}(r) = \frac{1}{2^{k}-1}\ round((2^{k}-1)\ r). 
\label{quantizer}  
\end{equation}
Due to the non-differentiability of the quantizer, the gradient $\frac{\partial q}{\partial r}$ is approximated as $1$ during back-propagation, namely Straight-Through-Estimator~\cite{hubara2016binarized}.

\subsection{Dynamic Inference}
Dynamic inference is the technique to flexibly adjust the network structure during inference to satisfy requirements of computing resources or different tasks. Models allowing dynamic inference can be viewed as an integration of a bunch of sub-DNNs. According to the dimension along which to integrate, existing integration models can be divided into three classes. Firstly, Slimmable DNNs~\cite{yu2018slimmable, yu2019universally} are a type of dynamic DNN that can execute at different channel widths, which can instantly adjust their memory footprint during inference. Then, Mult-Exit DNNs~\cite{li2019improved, phuong2019distillation} attach multiple classifiers to network structure at different layers, and can decide the model depth to make predictions for fast inference. Different from dynamic models mentioned above, Superposition~\cite{cheung2019superposition} integrates multiple sub-DNNs that are targeted at different tasks, each sub-DNNs can be retrieved during inference.

\section{Quantizable DNNs}

\subsection{Problem Statement}
We first formally define Quantizable DNN under supervised learning setting. Given a training data set $S_{t}=\{(x_{i}, y_{i})\}_{i=1}^{N}$, where $\{x_{i}\}_{i=1}^{N}$ are the input variables and 
$\{y_{i}\}_{i=1}^{N}$ are the corresponding target variables, a $k$-bit Quantizable DNN is trained as a special type of Quantized DNN $\hat{y}=\mathcal{F}_{k}(x; W_{k})$, where $\hat{y}$ is the prediction for the corresponding target variable, $k$ is the model's bit-width, $W_{k}$ is the quantized model weight in $k$-bit, $x$ is the input of one data sample. 
Different from existing $k$-bit quantized DNN which can only run with the fixed bit width and requires further re-training to change its bit-width at run time, the Quantized DNN can flexibly adjusts its bit-width, i.e. bit mode, on the fly. Given a desired bit mode $b (1\leq b \leq k)$ at run-time, the $k$-bit Quantizable DNN can be switched to $b$-bit mode with $\mathcal{F}_{b}(x; W_{b}) =Q(\mathcal{F}_{k}(x; W_{k}), b)$, where $Q$ is a pre-defined quantization function. For Quantizable DNN, we want to simultaneously maximize the accuracy of $\mathcal{F}_{b}(x; W_{b}), \forall b \in (1\leq b \leq k)$.

\subsection{Overall Framework}

We denote $\mathcal{F}_{k}(x; W_{k})$ as the $k$-bit Quantizable DNN. 
During training, a 32-bit mode is trained, which serves as the `parent' model for other bit modes, i.e. $mode\ list_{train}=\{1,2,...,k,32\}$. The `parent' model is jointly optimized by all bit modes in $mode\ list$ under multi-task frameworks. For a mini-batch of training data, the Quantizable DNN conducts forward and backward computations in each bit mode and accumulates the gradient. Weights are updated after traversing all bit modes. In the following section, we first introduce the \emph{Consistency-based Training Objective} to optimize the model, which enforces lower bit modes to produce consistent performance with 32-bit mode. To resolve the conflicts between different bit modes, we propose \emph{Bit-Specific Batch Normalization} to normalize outputs in different modes with corresponding learnable affine functions, which is crucial for the performance of Quantizable DNNs. The $k$-bit Quantizable DNN can be directly retrieved from the 32-bit `parent' model with pre-defined $k$-bit quantizer $Q_{k}(\cdot)$, i.e.  $\mathcal{F}_{k}(x; W_{k})=Q_{k}(\mathcal{F}_{32}(x; W_{32}))$. To switch to $b$-bit mode from $\mathcal{F}_{k}(x; W_{k})$, a special quantization function is needed, namely Switch function. We re-design the pre-defined quantizer via \emph{Thresholds Alignment} to ensure
the existence of switch function $S(\cdot)$.

\subsection{Consistency-based Training Objective}
In this section, we introduce the overall optimization objective of the Quantizable DNN. Since 32-bit mode is introduced as the `parent' model for training, we denote the Quantizable DNN as $\mathcal{F}_{32}(x,W_{32})$ accordingly. Training a Quantizable DNN can be formulated as a multi-task problem, where each bit mode is treated as a sub-task. The sub-loss function attached to $i$-bit mode is denoted as $\mathcal{L}_{i}$. Integrating these sub-loss functions, we can obtain the overall training objective $\mathcal{L}_{all}$:

\begin{equation}
\mathcal{L}_{all} = \sum_{k \in mode\ list}\alpha_{k} \mathcal{L}_{k} + \gamma||Q||^{2},
\label{lall} 
\end{equation}

where $\alpha_{k}$ denotes pre-defined weights, and $\gamma$ is the balancing hyper-parameter between empirical loss and regularization (weight decay). A higher value of $\alpha_{k}$ encourages the Quantizable DNN to put more attention on $k$ bit mode. In this paper, we treat each bit mode equally and set $\alpha_{k}=1$ for all $k$. In classification tasks, for $k=32$, $\mathcal{L}_{k}$ is simply the widely-adopted cross-entropy loss supervised by ground-truth $y$. However, for $k<32$, we instead introduce consistency loss to ensure their performance, which utilizes predictions of $32$ bit mode $\mathcal{F}_{32}(x, W_{32})$ as supervision for $\mathcal{F}_{k}(x, W_{k})$. Note that $\mathcal{F}_{k}(x, W_{k})$ can be directly re Such strategy is adopted because there should be certain internal consistency between predictions of different bit modes since they are integrated into a unified structure. The consistency loss is forumated as Eq.(\ref{lk}):

\begin{equation}
\mathcal{L}_{k (k<32)} = 
KL(\sigma(\frac{\mathcal{F}_{32}(x, W_{32})}{T}),\ \sigma(\frac{\mathcal{F}_{k}(x, W_{k})}{T})),
\label{lk} 
\end{equation}

where $KL(\cdot)$ and $\sigma(\cdot)$ denote Kullback-Leibler divergence and softmax function respectively. Inspired by Knowledge Distillation~\cite{hinton2015distilling}, we introduce a hyper-parameter $T$ to control the smoothness of the supervision, which can explore the `dark' knowledge between classes. Note that gradients from consistency loss to $\mathcal{F}_{32}(x,W_{32})$ are ignored.

\subsection{Bit-specific Batch Normalization}

Batch Normalization~\cite{ioffe2015batch} is proposed to normalize the channel-wise features $y$ with a set of parameters $(\gamma, \beta, \mu, \sigma)$:

\begin{equation}
BN(y) = \gamma \frac{y-\mu}{\sqrt{\sigma^{2} + \epsilon}} + \beta,
\label{bn} 
\end{equation}
where $\gamma$ and $\beta$ are learnable parameters, $\epsilon$ is a small value which can be neglected. $\mu$ and $\sigma^{2}$ are means and variances of channel-wise features. During training, $\mu$ and $\sigma^{2}$ are calculated by the current mini-batch. During evaluation, $\mu$ and $\sigma^{2}$ are moving average of all training set. Batch Normalization is crucial for quantized DNNs, which maps activations to approximate Gaussian distributions $\mathcal N (0, 1)$ to make most values lie in quantization interval. However, for Quantizable DNNs, since matrix multiplication outputs produced by different bit modes have different distributions, shared Batch Normalization cannot properly normalize outputs for all bit modes.

Instead of sharing Batch Normalization layers, we propose \emph{Bit-specific Batch Normalization}, which has two variants. For \textbf{variant A}, we assign private $(\mu_{k}, \sigma_{k})$ for each bit mode. When training or inference in different modes, channel-wise feature maps can be mapped to the same distribution with respective $(\mu_{k}, \sigma_{k})$ for further quantization.  Formally, we denote $y_{k}=conv(w_{k}, a_{k})$, where $a_{k}$ represents the k-bit activations from the previous layer and $w_{k}$ is the k-bit weights in the current convolution layer. $conv$ denotes convolution operation and $y_{k}$ is its output. Note that the output $y_{k}$ is not limited to k-bit. The \textbf{variant A} of \emph{Bit-specific Batch Normalization} can be defined as:

\begin{equation}
BSBN^{A}(y_{k}) = BSBN^{A}_{k}(y_{k}) =  \gamma \frac{y_{k}-\mu_{k}}{\sqrt{\sigma^{2}_{k} + \epsilon}} + \beta,
\label{bsbna} 
\end{equation}

where $\gamma$ and $\beta$ are shared learnable parameters. $\mu_{k}$ and $\sigma^{2}_{k}$ are private statistical parameter for $k$-bit mode, which can be either updated as other parameters or directly estimated during inference via post-training strategy~\cite{yu2019universally}. With such strategy, \textbf{variant A} can introduce no additional parameters and enable Quantizable DNNs to achieve usable performance. Based on \textbf{variant A}, we further introduce a \textbf{variant B}, which assigns not only private $(\mu_{k}, \sigma_{k})$, but also private $(\gamma_{k}, \beta_{k})$ to each mode: 

\begin{equation}
 BSBN^{B}_{k}(y_{k}) =  \gamma_{k} \frac{y_{k}-\mu_{k}}{\sqrt{\sigma^{2}_{k} + \epsilon}} + \beta_{k}.
\label{bsbnb} 
\end{equation}

The \textbf{variant B} can bring more flexibility to each bit mode and further ease their conflicts. For example, channels that have texture-rich information in higher bit modes may convey negligible information in low-bit modes, and the latter can assign a lower value to the corresponding $\gamma_{k}$ to reduce their interference. Though \textbf{variant B} introduces additional parameters, the cost can be neglected, because the parameters in Batch Normalization are usually less than 2\% of the total model. And it has no effect on inference speed, since in each mode, only the corresponding normalization operation is included in inference graph. \textbf{Variant B} is adopted in our experiments unless otherwise stated.

\subsection{Quantizer Re-design by Thresholds Alignment}

\begin{figure*}[ht]
\vskip 0.2in
\begin{center}
\centerline{\includegraphics[width=18cm]{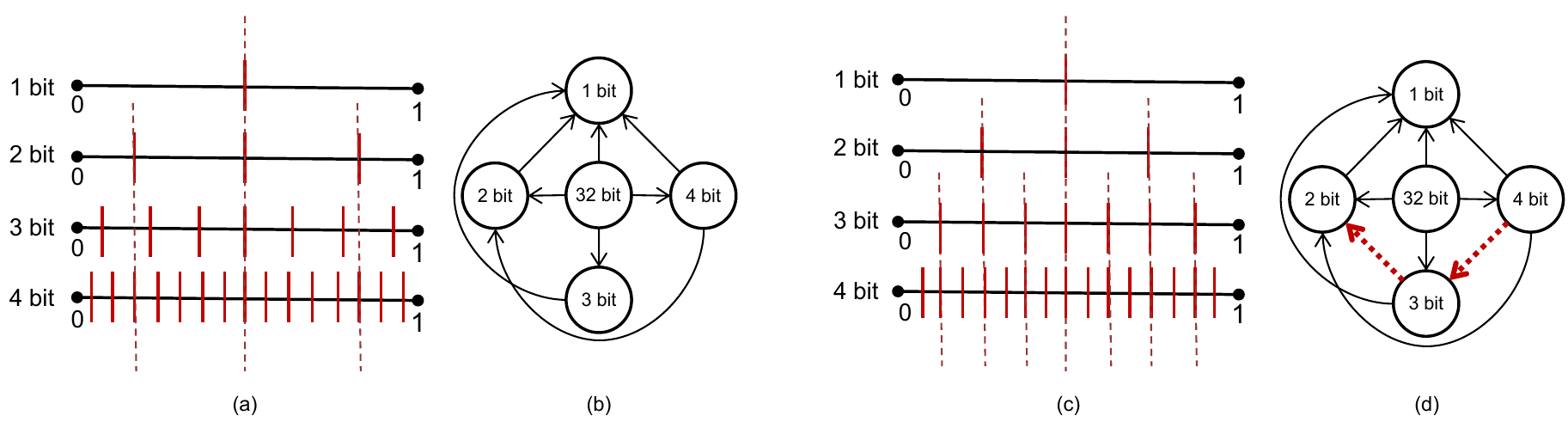}}
\caption{Comparision between Quantizable DNNs with (Subfigure \textbf{(c)} and \textbf{(d)}) and without (Subfigure \textbf{(a)} and \textbf{(b)}) thresholds alignment . }
\label{graph}
\end{center}
\vskip -0.2in
\end{figure*}

Equipped with components introduced above, $b$-bit modes can be directly retrieved from the 32-bit `parent' model with $b$-bit quantizer, i.e. $\mathcal{F}_{b}(x; W_{b})=Q_{b}(\mathcal{F}_{32}(x; W_{b}))$.
However, it may fail to retrieved from a $k$-bit Quantizable DNN, where $b<k<32$, because the Switch function $S(\cdot)$ that maps $k$-bit weights $W_{k}$ to $b$-bit weights $W_{b}$ may not exist. To solve this prolem, we re-design the pre-defined quantizer with `thresholds alignment' constraint. To be specific, the quantization thresholds in lower bit modes are forced to a subset of that in higher bit modes. Below we prove the necessity of this constraint.

Suppose we want to switch to $b$-bit mode with a $k$-bit Quantizable DNN, whose weights are stored in $k$-bit, where $i < k < 32$. The `real' $b$-bit weights $W_{b}$ are directly quantized from `parent' 32-bit weights, which can be represented as $W_{b}= Q_{b}(W_{32})$, where $Q_{b}(\cdot)$ is the $b$-bit pre-defined quantizer, $W_{32}$ is the weights of the `parent' model. To obtain `real' $b$-bit weights $W_{b}$ from only $k$-bit weights $W_{k}$, we assume there exists a Switch function $S(\cdot)$, which can perfectly map $W_{k}$ to desired $W_{b}$. According to the definition, the Switch function must satisfy:

\begin{equation}
W_{b} = Q_{b}(W_{32}) = S(W_{k},b) = S(Q_{k}(W_{32}),b).
\label{align1}  
\end{equation}

Rewrite Eq.(\ref{align1}) in element-wise form, then we can obtain:

\begin{equation}
Q_{b}(w_{32}^{n}) = S(Q_{k}(w_{32}^{n}),b), \  0 \leq n \leq N-1,
\label{align2}  
\end{equation}

where $N$ is the number of elements of weights. From Eq.(\ref{align2}), pre-defined quantizers $Q(\cdot)$ should satisfy that for any $0 \leq m < n \leq N-1$, if $Q_{b}(w_{32}^{m}) \neq  Q_{b}(w_{32}^{n})$, then $Q_{k}(w_{32}^{m}) \neq  Q_{k}(w_{32}^{n})$, where $b < k$. To achieve this, one requirement is that thresholds of $Q_{b}(\cdot)$ must be a subset of that of $Q_{k}(\cdot)$. Therefore, the constraint of thresholds alignment is necessary for the Switch function $S(\cdot)$ to exist. 

However, the widely-used quantizer defined in Eq.(\ref{quantizer}) does not satisfy such constraint. Since $\{\frac{2i+1}{2}|i \in Z\}$ are thresholds for $round(\cdot) $ function, thresholds $T^{k}$ for Eq.(\ref{quantizer}) can be written as: 

\begin{equation}
T_{k} = \{\frac{2i+1}{2^{k+1}-2}|\ 0 \leq i < 2^{k}-1, i \in Z \}.
\label{thresholds1}  
\end{equation}

For different bit-width $k$, thresholds $T_{k}$ are a bunch of fractions with the denominator of $2^{k+1}-2$, which obviously not always satisfy $T_{k-s}\subset T_{k}$, where $0<s<k$. To make pre-defined quantizers satisfy the constraint discussed above, we re-design the quantizer defined by Eq.(\ref{quantizer}) into Eq.(\ref{quantizer2}), i.e. thresholds alignment:  

\begin{equation}
\small{Q_{k}^{a}(r) = \frac{1}{2^{k}-1}clamp\ (round\ (2^{k} r -0.5),0,2^{k}-1)},
\label{quantizer2}  
\end{equation}

where $r$ and $k$ represent full-precision (32-bit) value and bit-width respectively. $clamp(\cdot)$ is applied to handle special cases, i.e. when $r=1$. We visualize thresholds of Eq.(\ref{quantizer}) and Eq.(\ref{quantizer2}) in Subfigure \textbf{(a)} and \textbf{(c)} Fig.~\ref{graph} respectively, where solid vertical lines represent thresholds of quantizers. 

With Eq.(\ref{quantizer2}), the Switch function $S(\cdot)$ that converts $k$-bit stored model to $b$ bit weights can be formulated as Eq.(\ref{switch}), where $b<k<32$.

\begin{equation}
\small{S(W_{k},b)= \frac{1}{2^{b}-1} round \left(\frac{2^{k}-1}{2^{k-b}}W_{k}-0.5\right)}
\label{switch}  
\end{equation}

We demonstrate the effectiveness of thresholds alignment in Subfigure \textbf{(b)} and \textbf{(d)} of Fig.~\ref{graph}, where nodes denote the bit-widths to store the Quantizable DNN and edges denote that it can switch from the highest bit mode to lower bit modes. For example, only with thresholds alignment, the $4$-bit stored Quantizable DNN can then be switched to $3$-bit mode, as is represented with dotted arrows in Subfigure \textbf{(d)}. In our experiments, thresholds alignment leads to no significant difference in prediction accuracy.

\subsection{Overall Training Algorithm}

Algorithm~\ref{alg:example} demonstrates the overall training algorithm of a $k$-bit Quantizable DNN, which can instantly adjust its bit-width from 1 to $k$ bit. During training, only the `parent' model $\mathcal{F}_{32}$ and the Bit-Specific Batch Normalizations $BSBN$ need to be updated. After training, we can obtain the $k$-bit Quantizable DNN $\mathcal{F}_{k}$ by quantizing 32-bit weights of $\mathcal{F}_{32}$. Due to the \emph{thresholds alignment} in line 4, each bit mode can be retrieved from higher bit modes rather than only the 32-bit parent. Therefore, the model can instantly execute in $b$-bit ($b<k$) mode $\mathcal{F}_{k}(x,i)$ with switch function $S(\cdot)$.

\begin{algorithm}%[tb]
   \caption{Training a $k$-bit Quantizable DNN}
   \label{alg:example}
\begin{algorithmic}[1]
   \STATE {\bfseries Define:} hyper-parameter $T$ in consistency loss.\\
   \STATE Initialize `parent' model $\mathcal{F}_{32}$.\\
   \STATE Initialize Bit-Specific Batch Normalizations $BSBN$.\\
   \STATE Re-design quantizers with thresholds alignment.\\
   \STATE \textbf{for} $i = 1,...,n_{iters}$ \textbf{do} \\
   \STATE \quad Get next mini-batch of training data $(x, y)$.\\
   \STATE \quad Clear gradients.\\
   \STATE \quad Switch $BSBN$ to $BSBN_{32}$.\\
   \STATE \quad Forward 32 bit mode, $y_{32} = \mathcal{F}_{32}(x,W_{32})$.\\
   \STATE \quad Compute Cross-entropy loss $\mathcal{L}_{32}(y,\ y_{32})$\\ 
   \STATE \quad Back-propagate to accumulate gradients.\\
   \STATE \quad Save $y_{32}$ for consistency loss.\\
   \STATE \quad \textbf{for} $i$ in \{1,2,...,$k$\} \textbf{do}\\
   \STATE \quad \quad Obtain $i$-bit weights $Q_{i}^{a}(W_{f})$.\\
   \STATE \quad \quad Switch $BSBN$ to $BSBN_{i}$.\\
   \STATE \quad \quad Wrap activation functions with $Q_{i}^{a}(\cdot)$.\\
   \STATE \quad \quad Forward $i$ bit mode $y_{i} = \mathcal{F}_{i}(x,W_{i})$.\\
   \STATE \quad \quad Compute Consistency loss $\mathcal{L}_{i}(y_{32},\ y_{i})$.\\ 
   \STATE \quad \quad Back-propagate to accumulate gradients.\\
   \STATE \quad \textbf{end for}\\
   \STATE \quad Update `parent' model $\mathcal{F}_{32}$ and BSBN.\\
   \STATE \textbf{end for}\\
   \STATE Quantize  $\mathcal{F}_{32}$ to $\mathcal{F}_{k}$ and store the model.\\
\end{algorithmic}
\end{algorithm}

\section{Experiments}

\subsection{Implement Details}
To validate the performance of Quantizable DNNs, we compare it with individual quantized DNNs on  Cifar100~\cite{krizhevsky2009learning} and ImageNet~\cite{deng2009imagenet} datasets, in terms of classification accuracy. 
Our implementation are based on PyTorch~\cite{paszke2019pytorch}. Cifar100 has 40,000 training images, 10,000 validation images and 10,000 test images. Note that since there is no official split, we divide training/validation set by ourselves. ImageNet has 1,280,000 training images and 50,000 validation images.  Results on CIFAR100 are average of $3$ runs. To ensure fairness, Both Quantizable DNNs and corresponding quantized DNNs are trained from scratch for the same epochs, with the same batch size and learning rate. Common data augmentation techniques, e.g. Random Resized Crop and Random Horizontal Flip are adopted for both models. The hyper-parameter $T$ for \emph{consistency loss} is empirically set to $2$ for all experiments, which is selected based on the validation set of CIFAR100.

\subsection{Classification Performance}
Table~\ref{cifar} and Table~\ref{imagenet} provide the results on CIFAR100 and ImageNet, respectively. 
On CIFAR100, we experiment with a Resnet variant which removes the first pooling layer due to the small image size (32$\times$32). All models in Table~\ref{cifar} are trained for 100 epochs with the batch size of 128. On ImageNet, Quantizable DNNs are implemented based on standard AlexNet and Resnet-18. All experiments are trained for 45 epochs with a batch size of 256.

It can be seen that a single Quantizable DNN can even achieve higher overall classification accuracy than a bunch of individual quantized DNNs. On CIFAR100, Low bit modes in Quantizable DNNs outperforms quantized DNNs by 2.81\% to 3.29\% . And 32-bit mode achieves 0.71\% gains due to the regularization from lower bit modes. Similar results are also observed for Alexnet on ImageNet. The accuracy gains verify the effectiveness of the co-regularization scheme in Quantizable DNNs. When experiment with Resnet-18 on ImageNet, the performance of $32$-bit mode is degraded by $-1.33\%$, we attribute it to its compact network architecture, which is more likely to be over-regularized. Note that $1$-bit mode is not included in mode list for Quantizable Resnet-18 due to its unique incompatibility, which will be further explained in \textbf{Section 4.3}.

\begin{table*}[]
\centering
\caption{Top-1 test accuracy on CIFAR100.}\smallskip
%\resizebox{0.95\textwidth}{!}{ % If your table exceeds the column or page width, use this command to reduce it slightly
\begin{tabular}{lclcc}
  \toprule
 Model & Top1 Accuracy(\%) & Model & Top1 Accuracy & Accuracy gain(\%)\\
  \midrule
  32-bit Resnet-Cifar  &  70.66 $\pm$ \ 0.16 &  & 71.37 $\pm$ \ 0.25 & +0.71 $\pm$ \ 0.30\\
  4-bit Quantized Resnet-Cifar & 68.26 $\pm$ \ 0.12 & & 71.25 $\pm$ \ 0.17 & +2.99 $\pm$ \ 0.21 \\
  3-bit Quantized Resnet-Cifar & 67.87 $\pm$ \ 0.19 & Quantizable Resnet-Cifar & 71.16 $\pm$ \ 0.29 & +3.29 $\pm$ \ 0.35  \\
  2-bit Quantized Resnet-Cifar & 67.69 $\pm$ \ 0.26 & & 70.50 $\pm$ \ 0.59 & +2.81 $\pm$ \ 0.64\\
  1-bit Quantized Resnet-Cifar & 61.92 $\pm$ \ 0.30 & & 64.95 $\pm$ \ 0.16 & +3.03 $\pm$ \ 0.34\\
  \bottomrule
 \end{tabular}
\label{cifar}
\end{table*}

\begin{table*}[]
\centering
\caption{Top-1 validation accuracy on ImageNet.}\smallskip
%\resizebox{0.95\textwidth}{!}{ % If your table exceeds the column or page width, use this command to reduce it slightly
\begin{tabular}{lclcc}
  \toprule
  Model & Top1 Accuracy(\%) & Model & Top1 Accuracy & Accuracy gain(\%)\\
  \midrule
  32-bit Alexnet   &  61.38\% &  & 62.86\% &+1.48\%\\
  4-bit Quantized Alexnet & 60.67\% & & 61.68\% & +1.01\% \\
 3-bit Quantized Alexnet & 58.88\% & Quantizable Alexnet & 60.76\% & +1.88\% \\
   2-bit Quantized Alexnet & 52.58\% & & 56.66\% & +4.08\%\\
  1-bit Quantized Alexnet & 38.97\% & & 39.98\% & +1.01\%\\
 % \cmidrule(lr){2-2}\cmidrule(lr){3-3}\cmidrule(lr){4-4}\cmi%drule(lr){5-5}\cmidrule(lr){6-6}
  %ImageNet & AVG & 54.50\% & AVG & 56.39\% & $\bm{+1.89\%}$\\
  \midrule
  32-bit Resnet-18   &  68.60\% &  & 67.27\% & -1.33\% \\
  4-bit Quantized Resnet-18 & 65.93\% & & 66.94\% & +1.01\% \\
  3-bit Quantized Resnet-18 & 65.03\% & Quantizable Resnet-18 & 66.28\% & +1.25\% \\
  2-bit Quantized Resnet-18 & 61.73\% & & 62.91\% & +1.18\%\\
  1-bit Quantized Resnet-18* & 50.67\% & & - & -\\
 % \cmidrule(lr){2-2}\cmidrule(lr){3-3}\cmidrule(lr){4-4}\cmid%rule(lr){5-5}\cmidrule(lr){6-6}
 %& AVG & 65.32\% & AVG & 65.85\% & $\bm{+0.53\%}$\\
  \bottomrule
 \end{tabular}
\label{imagenet}
\end{table*}

\subsection{Analysis of Aggressive 1-Bit Mode}

\begin{figure*}[t]
\vskip 0.2in
\begin{center}
\centerline{\includegraphics[width=17.5cm]{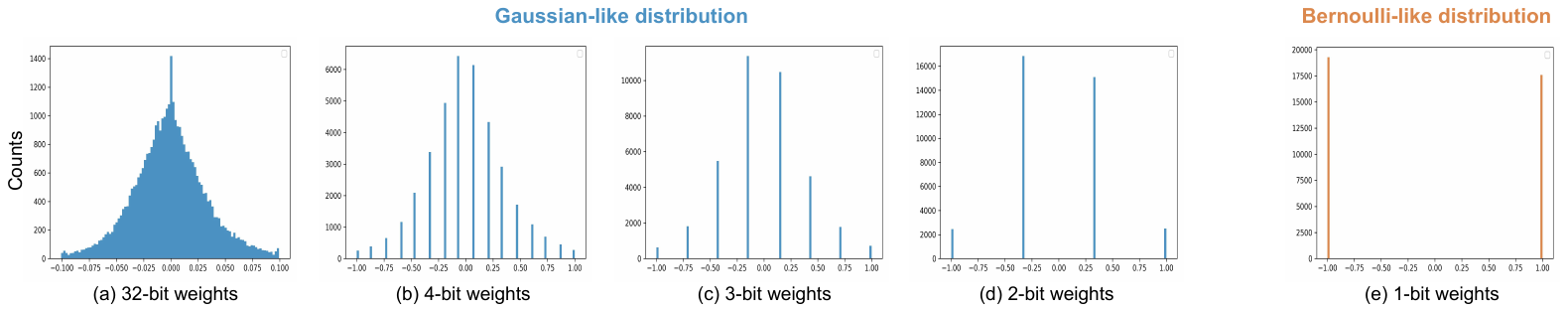}}
\caption{Visualization of weight distributions in different bit modes. 1 bit mode has fundamental difference with other modes.}
\label{gaussian}
\end{center}
\vskip -0.2in
\end{figure*}

When conducting experiments under (ImageNet, Resnet-18) setting, we observe that the performance of 1-bit mode is significantly worse than 1-bit quantized Resnet-18 (by $\approx -7\%$). In response to this phenomenon, we conduct a special analysis for 1-bit mode in this section. Compared to 2-4 bit modes, the most notable feature of 1-bit mode is the mutation of distribution characteristics, which is demonstrated in Fig.~\ref{gaussian}. When quantized to 2-4 bit, the quantized weights still hold a Gaussian-like (bell-shaped) distribution. However, when further quantized to 1-bit, weights then turn into Bernoulli distribution, which may make it difficult for 1-bit mode to be compatible with other bit modes when integarted into a unified model.

\begin{figure}[ht]

\begin{center}
\centerline{\includegraphics[width=9cm]{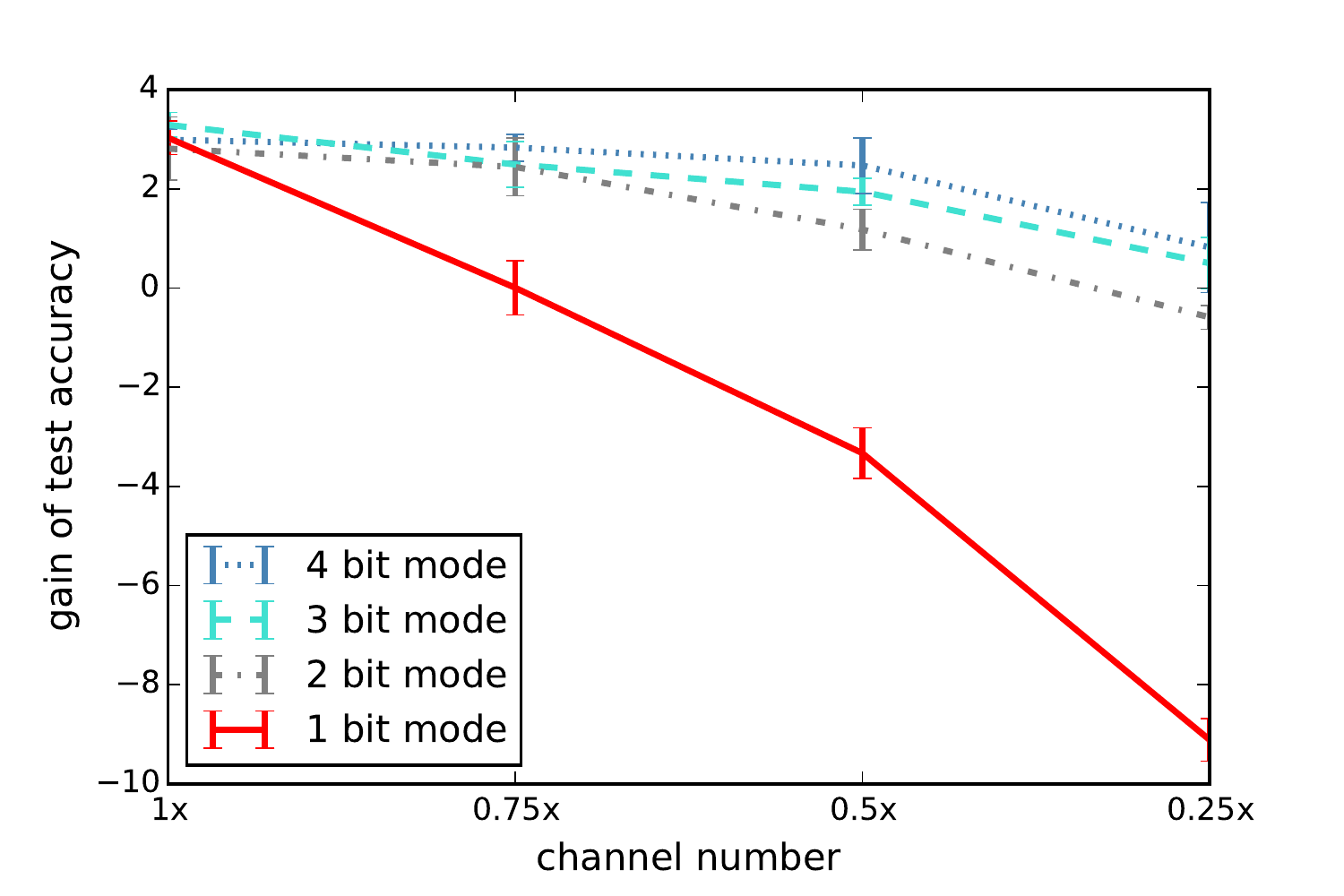}}
\caption{Gain of test accuracy for different bit modes on models with different channel numbers.}
\label{1bit}
\end{center}
\vskip -0.2in
\end{figure}

However, for experiments under (ImageNet, Alexnet) and (CIFAR100, Resnet) settings, no degradation is observed for 1-bit mode. We conjecture it is the redundant capacity that allows the model to tolerate the incompatibility from 1-bit mode, because Alexnet (239M) has much more parameters than Resnet-18 (46M) and the task of CIFAR100 is much easier than ImageNet. To validate this hypothesis, we conduct experiments on CIFAR100 using Quantizable Resnet-Cifar with different channel numbers ($1\times, 0.75\times, 0.5\times, 0.25\times$). The fewer channels there are, the less redundancy there is in the model. Fig.~\ref{1bit} shows the accuracy gains compared with quantized DNNs. It can be seen that as the channel number decreases, performance gains of 1-bit mode decreases rapidly compared with 2-4 bit modes. This phenomenon reveals that the more compact the model is, the more obvious the incompatibility from 1-bit mode is. On this basis, we further speculate that 1-bit mode can bring negative impacts on other bit modes when integrated in a compact model, and Table~\ref{train1bit} verifies our speculation. Therefore, the 1-bit mode is discarded for Quantizable Resnet-18 (Table~\ref{imagenet}).

\begin{table}[htpb]
\centering
\caption{Impacts of 1-bit mode on other bit modes. `Baseline' denotes quantized Resnet-Cifar with 0.25$\times$ channel number. Model A and Model B are Quantizable Resnet-Cifar with and without 1-bit mode respectively. }\smallskip
%\resizebox{0.95\textwidth}{!}{ % If your table exceeds the column or page width, use this command to reduce it slightly
\begin{tabular}{lccc}
  \toprule
  Mode & Baseline (\%) & Model A (\%) & Model B (\%) \\
  \midrule
  32-bit  &   63.36  $\pm$\ 0.29   & 61.88  $\pm$\ 0.77 & 63.73  $\pm$\ 0.34\\
  4-bit  &  61.20  $\pm$\ 0.61 & 62.03  $\pm$\ 0.67 & 63.60  $\pm$\ 0.41\\
  3-bit &  60.81  $\pm$\ 0.29  & 61.32  $\pm$\ 0.44 & 63.01  $\pm$\ 0.24\\
  2-bit  &  58.58  $\pm$\ 0.20 & 57.99  $\pm$\ 0.13 & 58.34  $\pm$\ 0.10 \\
  1-bit &  46.49  $\pm$\ 0.37 & 37.37  $\pm$\ 0.22 & - \\
  \bottomrule
 \end{tabular}
\label{train1bit}
\end{table}

\subsection{Ablation Study}

We propose \emph{Bit-Specific Batch Normalization} to enable Quantizable DNNs to converge, and optimize it with \emph{Weighted Consistency-based Training Objective}. To verify the effectiveness of these two components, ablation study is conducted on CIFAR100.

\subsubsection{Effectiveness of Bit-specific BN}

In this section, we make a comparision between Batch Normalization~\cite{ioffe2015batch}, \emph{Bit-Specific Batch Normalization} varaint A and varaint B. Results are presented in Table~\ref{sharebn}. Since BN fails to resolve the conflict between different bit modes, it produces poor overall performance as expected. For BSBN$^{A}$, since we normalize feature maps from different bit modes with private $(\mu_{k}, \sigma_{k})$, significant improvement is observed for all bit modes. It futher verifies that the differences in output distribution is the main conflict among different bit modes. On this basis, BSBN$^{B}$ further assigns more flexibility to each bit mode via private $(\gamma_{k}, \beta_{k}, \mu_{k}, \sigma_{k})$ and brings more performance gains to the model. In different situations, we can choose to use a particular variant as needed.

\begin{table}[htpb]
\centering
\caption{Ablation study for \emph{Bit-Specific Batch Normalization}. BN denotes Batch Normalization~\cite{ioffe2015batch}.  BSBN$^{A}$ and BSBN$^{B}$ are two variants of \emph{Bit-Specific Batch Normalization}.}
%\smallskip
%\resizebox{0.95\textwidth}{!}{ % If your table exceeds the column or page width, use this command to reduce it slightly
\begin{tabular}{lccc}
  \toprule
  Mode & BN(\%) & BSBN$^{A}$(\%) & BSBN$^{B}$(\%) \\
  \midrule
  32-bit  & 1.10  $\pm$\ 0.17  &  70.04  $\pm$\ 0.32   & 71.37  $\pm$\ 0.25 \\
  4-bit  & 6.36  $\pm$\ 1.59  & 69.73  $\pm$\ 0.37 & 71.25  $\pm$\ 0.17 \\
  3-bit & 20.99  $\pm$\ 3.45  & 69.84  $\pm$\ 0.35  & 71.16  $\pm$\ 0.29 \\
  2-bit  & 62.38  $\pm$\ 0.68  & 69.39  $\pm$\ 0.54 & 70.50  $\pm$\ 0.59 \\
  1-bit & 2.95  $\pm$\ 0.28  & 64.54  $\pm$\ 0.11 & 64.95  $\pm$\ 0.16 \\

  \bottomrule
 \end{tabular}
\label{sharebn}
\end{table}

\begin{table}[htpb]
\centering
\caption{Ablation study for \emph{Consistency-based Training Objective}.}\smallskip
%\resizebox{0.95\textwidth}{!}{ % If your table exceeds the column or page width, use this command to reduce it slightly
\begin{tabular}{lcc}
  \toprule
  Mode & Cross-entropy(\%) & Consistency loss(\%)\\
  \midrule
  32-bit  &   70.02  $\pm$\ 0.29   & 71.37  $\pm$\ 0.25 \\
  4-bit  &  69.75  $\pm$\ 0.31 & 71.25  $\pm$\ 0.17 \\
  3-bit &  69.63  $\pm$\ 0.07  & 71.16  $\pm$\ 0.29 \\
  2-bit  &  68.50  $\pm$\ 0.18 & 70.50  $\pm$\ 0.59 \\
  1-bit &  61.01  $\pm$\ 0.23 & 64.95  $\pm$\ 0.16 \\
  \bottomrule
 \end{tabular}
\label{loss}
\end{table}

\subsubsection{Effectiveness of Training Objective}
During training, each low bit mode is attached with a consistency loss rather than widely-used cross-entropy loss. The consistency loss enforces Quantizable DNNs to produce consistent predictions when degraded to low bit modes. Comparison between these two loss functions is presented in Table~\ref{loss}, and the consistency loss brings significant accuracy gains for all bit modes.

\section{Efficiency Analysis}
Quantizable DNNs can be viewd as the integration of multiple quantized DNNs. In this section we make a further comparision between quantized DNNs and Quantizable DNNs from the following aspects.

\paragraph{Training time}
The time occupied by data reading and augmentation cannot be ignored during training. To train multiple quantized DNNs with different bit-widths, we have to repeat the data-related process for multiple times. However, for Quantizable DNNs, less time is taken for training since all bit modes share the same training data. In our implement (GTX 1080 Ti, 20 Cpu cores), compared with the total time-consuming of individually trained Quantized Alexnets, Quantizable Alexnet consumes $0.9\times$ training time per epoch.

\paragraph{Memory footprint \& Inference speed}
During training/inference, the Quantizable DNN is converted from one bit mode to another, where each mode is equivalent to a quantized DNN. Therefore, Quantizable DNNs have the same memory footprint as corresponding quantized DNNs. For the same reason, both models can conduct inference at exactly the same high speed compared with 32 bit models.

\paragraph{Model size}
For $m$ quantized DNNs with different bit-widths $\{Q_{1}, \cdots, Q_{m}\}$, the total model size is $\sum\limits_{i=1}^{m} {model\ size(Q_{i})}$. When these m quantized DNNs are integrated into a single Quantizable DNN, the total model size is only slightly more than $\max\limits_{i}{model\ size(Q_{i})}$. Therefore, Quantizable DNNs take up less storage space than a bunch of corresponding quantized DNNs.

\section{Conclusion}
We propose the first DNN model that can adjust its bit-width on the fly, namely Quantizable DNNs. Compared with quantized DNNs, Quantizable DNNs can be instantly converted to different bit modes as needed, which provides much more flexibility in real application scenarios. Besides, the proposed model can even achieve higher accuracy than individual quantized DNN due to the co-regularization effects between 32-bit mode and low-bit mode. In the future, Quantizable DNNs can be combined with AutoML~\cite{he2018amc} to efficiently explore optimal bit-width for different layers.

\bibliography{example_paper}
\bibliographystyle{icml2020}

\end{document}